\documentclass[conference]{IEEEtran}
\usepackage{cite}
\usepackage{amsmath,amssymb,amsfonts}
\usepackage{graphicx}
\usepackage{textcomp}
\usepackage{xcolor}
\usepackage{bm}
\usepackage{amsmath}
\usepackage{amssymb}
\usepackage{amsthm}
\usepackage{mathrsfs}
\usepackage{enumerate}
\usepackage{multirow}
\usepackage{color}
\usepackage{threeparttable}
\usepackage{subfigure}

\usepackage[square, comma, sort&compress, numbers]{natbib}
\usepackage[pagebackref=false,breaklinks=true,letterpaper=true,colorlinks,bookmarks=false]{hyperref}
\usepackage{times}
\usepackage{breakurl}
\usepackage{array}
\usepackage{verbatim}
\usepackage{algorithm}
\usepackage{bbding}
\usepackage{algpseudocode}
\usepackage{lettrine}
\usepackage{booktabs}
\usepackage{tabularx}
\usepackage{adjustbox}
\usepackage{array} 
\usepackage{float}

\def\BibTeX{{\rm B\kern-.05em{\sc i\kern-.025em b}\kern-.08em
    T\kern-.1667em\lower.7ex\hbox{E}\kern-.125emX}}
\begin{document}

\title{Mirror Target YOLO: An Improved YOLOv8 Method with Indirect Vision for Heritage Buildings Fire Detection\\}

\author{\IEEEauthorblockN{1\textsuperscript{st} Jian Liang}
\IEEEauthorblockA{\textit{ College of Mechanical and Vehicle Engineering} \\
\textit{Hunan University}\\
Changsha City, China \\
hudajxgc@hnu.edu.cn}

\and
\IEEEauthorblockN{2\textsuperscript{nd} JunSheng Cheng*}
\IEEEauthorblockA{\textit{College of Mechanical and Vehicle Engineering} \\
\textit{Hunan University}\\
Changsha City,China \\
ChengJunSheng@hnu.edu.cn}
}

\maketitle

\begin{abstract}

Fires can cause severe damage to heritage buildings, making timely fire detection essential. Traditional dense cabling and drilling can harm these structures, so reducing the number of cameras to minimize such impact is challenging. Additionally, avoiding false alarms due to noise sensitivity and preserving the expertise of managers in fire-prone areas is crucial. To address these needs, we propose a fire detection method based on indirect vision, called Mirror Target YOLO (MITA-YOLO). MITA-YOLO integrates indirect vision deployment and an enhanced detection module. It uses mirror angles to achieve indirect views, solving issues with limited visibility in irregular spaces and aligning each indirect view with the target monitoring area. The Target-Mask module is designed to automatically identify and isolate the indirect vision areas in each image, filtering out non-target areas. This enables the model to inherit managers’ expertise in assessing fire-risk zones, improving focus and resistance to interference in fire detection.
In our experiments, we created an 800-image fire dataset with indirect vision. Results show that MITA-YOLO significantly reduces camera requirements while achieving superior detection performance compared to other mainstream models.

\end{abstract}


\section{Introduction}
Cultural relic buildings are valuable carriers of local history and culture ~\cite{ribera2020multicriteria}. Fires in these buildings pose a severe risk, as they could lead to irreplaceable losses~\cite{garcia2023fire}. Many memorials and museums fall under this category, yet, to preserve their historical integrity, they often lack conventional fire safety installations like sprinklers or smoke detectors~\cite{romao2022risk}. In the digital age, integrating deep learning algorithms for fire detection has emerged as a potential solution~\cite{dong2021survey}. However, due to structural constraints and obstructions within these buildings, effective camera coverage is challenging, requiring numerous cameras and extensive cabling, which risks damaging the building’s fabric.

\begin{figure*}[t]
    \centering
    \includegraphics[width=1\textwidth]{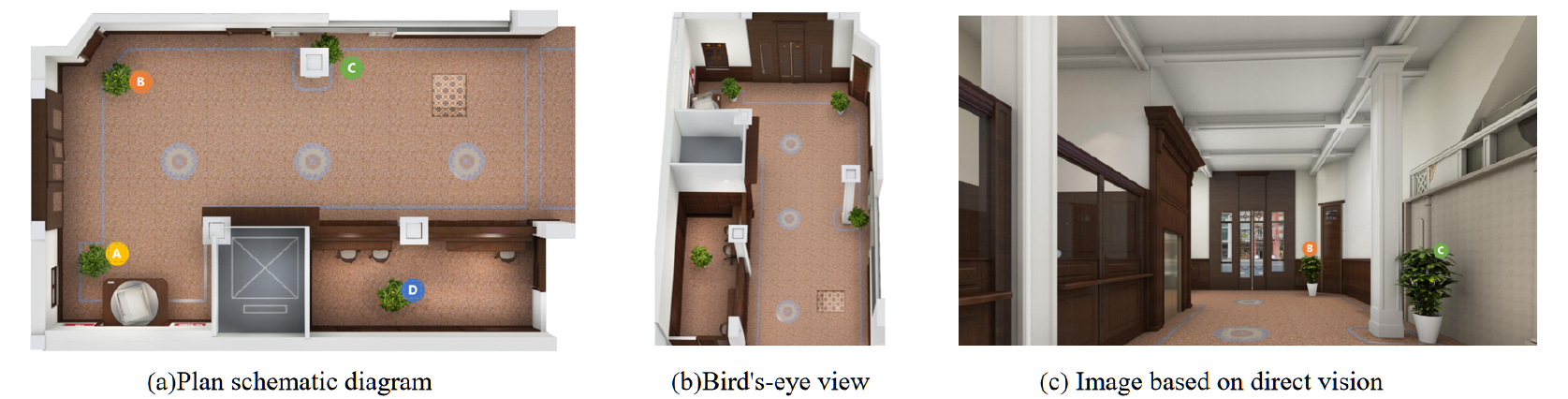}
    \caption{When the camera is deployed at a position directly facing the gate and only direct vision is used, due to the irregular shape of the indoor space, the direct vision of the camera is blocked. Among the four plants placed in this space, only two can be observed, and two plants are invisible in the occluded area.
}
    \label{fig:plan}
\end{figure*}

To address these challenges, a deep learning-based detection method is needed to expand each camera's monitoring range, ensure high detection accuracy, and minimize false alarms by filtering out low-risk areas. This approach would enable effective fire monitoring while preserving the historical integrity of cultural buildings.
Traditional fire monitoring methods and deep learning-based solutions represent two primary approaches. Conventional fire prevention relies on smoke alarms, temperature sensors, and manual fire alarm buttons~\cite{chen2022research}. Smoke alarms detect smoke particle concentration and issue alarms when a threshold is reached. Temperature sensors monitor environmental temperature fluctuations, signaling alarms when temperatures exceed safe levels. Manual fire alarm buttons allow individuals to alert the fire control center when a fire is observed.
However, installing smoke alarms and temperature sensors in architectural relics can significantly impact the building's structure and appearance. These installations often require extensive drilling and wiring, which risks structural damage and compromises the aesthetic and historical integrity of these sites. Thus, a more refined approach that reduces physical intervention while maintaining effective fire monitoring is essential for safeguarding cultural heritage.

Conventional deep learning fire prevention solutions use video surveillance combined with traditional deep learning algorithms, which significantly reduce the types and quantities of equipment needed, lessening the impact on the structure and historical features of cultural buildings~\cite{valikhujaev2020automatic,seydi2022fire,abdusalomov2021improvement}. However, these methods rely solely on direct vision, limiting the field of view per camera. In irregularly shaped spaces, multiple cameras are often required, which raises costs and still demands a lot of wiring and drilling, impacting the preservation and aesthetic value of these buildings.

To mitigate these issues, we propose a targeted fire prevention detection method based on indirect vision, named Mirror Target YOLO (MITA-YOLO). MITA-YOLO employs indirect vision through mirrors alongside an enhanced detection module. The approach uses wide-angle mirrors to extend the camera’s field of view, allowing monitoring of occluded areas without adding additional cameras. By strategically placing mirrors in areas with high fire risk (targeted detection areas) while avoiding low-risk zones (non-interest areas), this setup achieves focused surveillance without disrupting the structure.
In MITA-YOLO, the indirect vision generated by wide-angle mirrors effectively visualizes occluded regions, greatly reducing the number of cameras required in irregular spaces and avoiding structural damage from extensive wiring. By adjusting the mirror arrangement to focus only on high-risk areas, we ensure that each mirror image contains only targeted detection zones, filtering out irrelevant regions.
The proposed Target-Mask module enhances this capability by segmenting the indirect vision area in each image. Through a pre-trained model, Target-Mask automatically identifies mirror boundaries and uses these boundaries to restrict fire detection exclusively to targeted areas, filtering out non-interest regions~\cite{guo2022attention}. Consequently, MITA-YOLO focuses solely on relevant targets within the field of interest, improving detection accuracy and reducing false alarms This targeted approach achieves higher detection precision with lower missed detection rates, ensuring robust fire monitoring while preserving the historical integrity of cultural buildings.

\begin{figure*}[t]
    \centering
    \includegraphics[width=\textwidth]{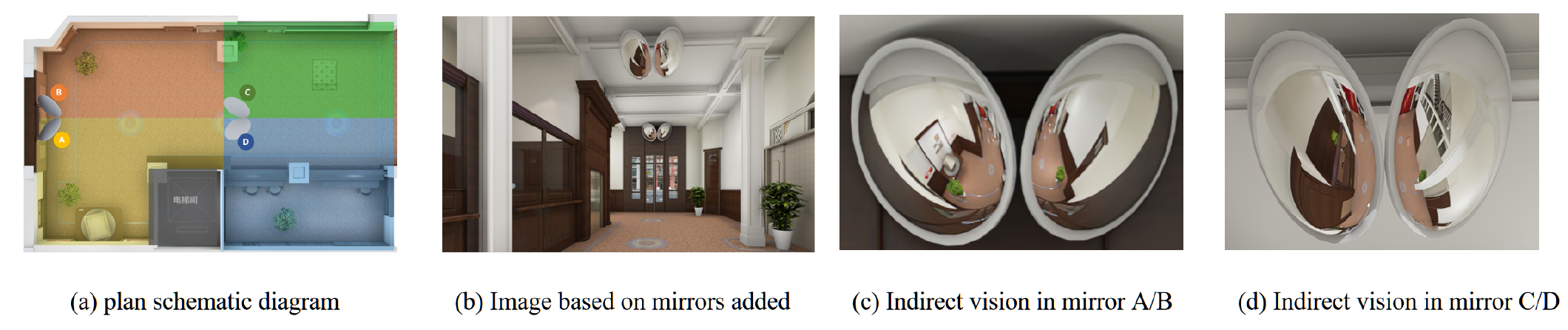}
    \caption{After the mirrors were deployed, the corresponding target areas could be seen in the indirect vision in the mirrors, realizing the expansion of the camera's field of vision coverage and the alignment of the indirect vision and the target detection area.}
    \label{fig:vision-text}
\end{figure*}

The main contributions of this paper are summarized as follows:

\begin{itemize} 
\item [\textcolor{black}{$\bullet$}] We integrate indirect vision technology into deep learning-based fire detection, leveraging mirrors to expand the camera's field of view, reduce camera deployment in irregular spaces, and align indirect vision with target detection zones. 
\item [\textcolor{black}{$\bullet$}] We propose a novel Target-Mask module that automatically identifies indirect vision boundaries and filters out irrelevant areas, enhancing detection accuracy and robustness against noise interference. 
\item [\textcolor{black}{$\bullet$}] To evaluate our approach, we simulate real-world scenarios in a cultural heritage building with irregular spaces, constructing a unique fire detection dataset consisting of 800 images captured via indirect vision. \item [\textcolor{black}{$\bullet$}] Our comparative experiments with YOLOv8n~\cite{terven2023comprehensive} demonstrate the effectiveness of our method, achieving a 3.7\% improvement in mAP50 and a 3\% increase in recall on the created dataset. \item [\textcolor{black}{$\bullet$}] MITA-YOLO achieves state-of-the-art performance in fire detection with indirect vision, outperforming six advanced deep learning models on our dataset. 
\end{itemize}

\section{Related Work}\label{sec:rw}

\subsection{Fire Detection in Heritage Buildings} 
Heritage buildings are invaluable legacies of human history and culture, holding significant historical and cultural value. They are not only the cornerstone of cultural heritage but also vital resources for promoting tourism and economic development. Protecting these buildings is a shared responsibility~\cite{Shan2022Investigating}, with fire prevention and detection identified as critical aspects of their preservation~\cite{Zhou2024Review}. Effective fire detection systems must address potential hazards without compromising the original structure or historical appearance of these buildings~\cite{tejedor2022non}.

Traditional fire prevention methods, such as smoke alarms and temperature sensors, have been widely used~\cite{khan2022recent}. However, these approaches often suffer from delayed detection, slow response times, and invasive installation requirements that can damage heritage structures~\cite{Zhang2018}. With advancements in deep learning, computer vision has demonstrated efficiency and accuracy in fire detection for conventional buildings~\cite{Zhang2020Integrating, Chaoxia2020, Liu2023}. 
To further enhance fire detection capabilities, diffusion models\cite{shen2023advancing,shen2024boosting} offer promising potential due to their ability to generate high-quality feature representations and model complex data distributions. 
Despite these advancements, current systems remain inadequate for the unique challenges posed by heritage buildings, such as maintaining structural integrity and minimizing aesthetic impact~\cite{Talaat2023Improved}.

To address these challenges, we propose MITA-YOLO, a targeted fire detection method tailored for heritage buildings. Our approach significantly reduces the number of cameras, wiring, and pipe installations while enhancing detection accuracy and minimizing missed detections. MITA-YOLO ensures effective fire prevention while preserving the historical and architectural integrity of heritage buildings.

\subsection{Detection Method Based on Indirect Vision} 
Indirect vision, commonly used in fields such as medicine, industry, and transportation, has garnered significant attention for its ability to extend the field of view and observe occluded targets. Chu et al.\cite{Chu2023Mirror} demonstrated its application in medicine, highlighting how adjusting mirror angles can align target areas for better observation. Similarly, Yan\cite{yan2024enhancing} explored its use in industrial scenarios, such as assisting with lane changes and reversing in vehicles. 
TCRL \cite{shen2023triplet} proposes a contrastive learning method that facilitates the interaction of global and local features to enhance the semantic saliency.
Shahar et al.~\cite{Shahar2012Attending} further noted the role of rearview mirrors in enabling drivers to monitor areas outside their direct line of sight. However, while these studies emphasize indirect vision's role in aiding human perception, its application in assisting computer vision remains underexplored.
MFC \cite{qiao2022novel} proposes a frequency-domain filtering module to achieve dense target feature enhancement.
CFIL \cite{weng2023novel} proposes a frequency-domain feature extraction module and feature interaction in the frequency domain to enhance salient features.
Existing fire detection models~\cite{Abdusalomov2023Improved, Zhao2022, Majid2022Attention} have achieved improvements in object detection, particularly for small targets, but fail to utilize the inherent benefits of mirror-based indirect vision. Traditional approaches often conduct full-image detection, introducing significant noise from irrelevant areas outside the mirrored regions.

To address these limitations, MITA-YOLO introduces a novel target-mask module that segments and isolates indirect vision areas in images. This module automatically identifies mirror edges, defines them as detection boundaries, and filters out irrelevant regions. By leveraging the pre-aligned target areas of indirect vision, the method minimizes noise interference and enhances detection accuracy. Furthermore, MITA-YO is compatible with various advanced object detection models, offering versatility and extensibility for broader applications.

\section{Proposed Method}\label{sec:method}
\begin{figure*}[t]
    \centering
    \includegraphics[width=\textwidth]{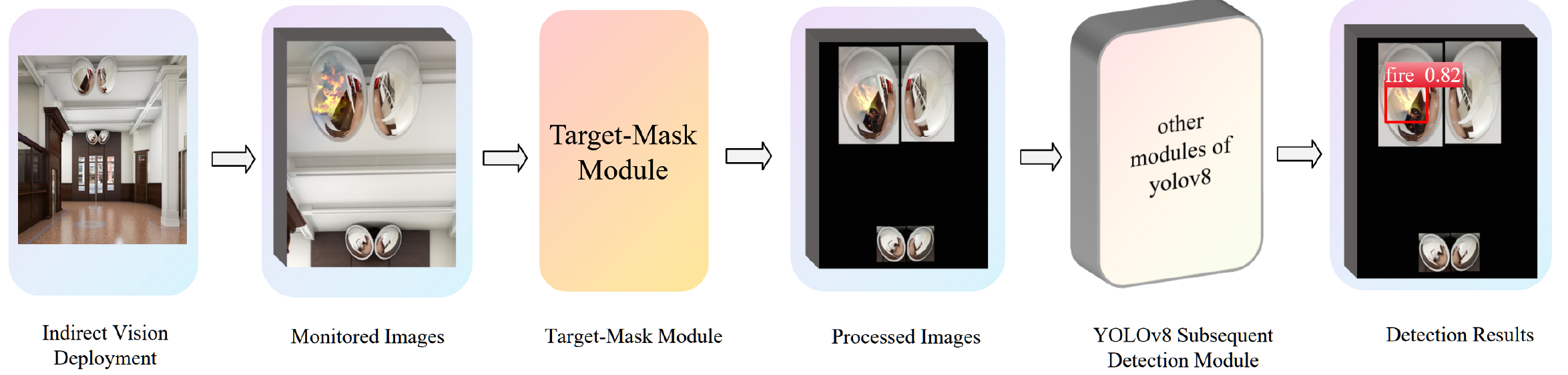}
    \caption{First, deploy the mirror system in the irregular space. Then, input the detection images obtained by the camera into the improved detection model. After the original images pass through the Target-Mask module, only the indirect vision area is retained. The processed images are then pushed into the subsequent modules of the detection model. Finally, the results of targeted detection only for the indirect vision area are obtained.}
    \label{fig:MITAYOLO}
\end{figure*}

The deployment of indirect vision in the interior space of cultural relic buildings and the addition of the Target-Mask module in the detection network are the improvement points proposed in this paper. Figure \ref{fig:MITAYOLO} illustrates the structure of the updated algorithm model.

\subsection{Deployment of Indirect Vision}
Due to the limitations of various obstructions and the irregular shape of the building layout, as well as the influence of various obstructions and changes in observation angles, the observation range of the camera through direct vision is greatly limited, and the monitoring range of each camera is extremely limited. Arranging a large number of cameras to achieve effective prevention and control coverage will bring large-scale pipe laying and wiring, which is very likely to cause irreversible damage to cultural relic buildings. At the same time, it is hoped that there is a more direct method to distinguish between targeted target areas and non-targeted areas to improve detection accuracy and reduce false alarms.

In order to solve the above problems, by reasonably arranging mirrors in irregular building spaces, the field of view of the camera is expanded through the indirect vision in the mirrors. Specifically, we use the reflection characteristics of mirrors here. It can change the observer's perspective without changing the position of the object. The indirect vision generated by it can provide people with more information. By adjusting the position and angle of the wide-angle mirror, the problem of blocked sight in an irregular space is solved, and the number of cameras required is also greatly reduced. At the same time, by adjusting the mirror to only observe the target monitoring area, the alignment of the indirect vision and the target monitoring area is achieved, that is, the indirect vision does not contain non-interest areas. At the same time, by adjusting the angle and focal length of the camera, the best image containing all indirect visions can be obtained.

For safety considerations, our mirrors can use acrylic mirrors. At the same time, the mirrors can be easily clamped on the building to avoid excessive nailing and fixing. At the same time, the significant reduction of cameras also reduces the corresponding amount of wiring and piping work. Reduces the probability of damage to cultural relic buildings.

\subsection{Target-Mask Module}  

There are many interference factors in fire scene detection. For example, the surrounding environment may present a complex scene with various background elements such as many decorations and crowds, which will introduce a large amount of redundant information and interference in the image. Due to the excessive repetition and interference brought by these background features, it will be challenging to distinguish fire and smoke targets from complex backgrounds. The basic principle of the Target-Mask module is to filter out irrelevant feature information with the cooperation of indirect vision and enhance useful feature information at the same time, so that the model can pay more attention to important areas in the image in a more adaptive way. Traditional attention mechanisms, such as CBAM (Convolutional Block Attention Module)~\cite{CBMA2018}, emphasize important channels and spatial positions in feature maps to improve the expressive ability of convolutional neural networks. CA (Channel Attention)~\cite{CA2021} focuses on the channel dimension of feature maps to achieve the strengthening of important channels, but they are all based on direct vision and do not consider the coexistence of indirect vision and direct vision and how to deal with their corresponding relationship.

\begin{figure}[ht]
    \centering
    \includegraphics[width=\linewidth]{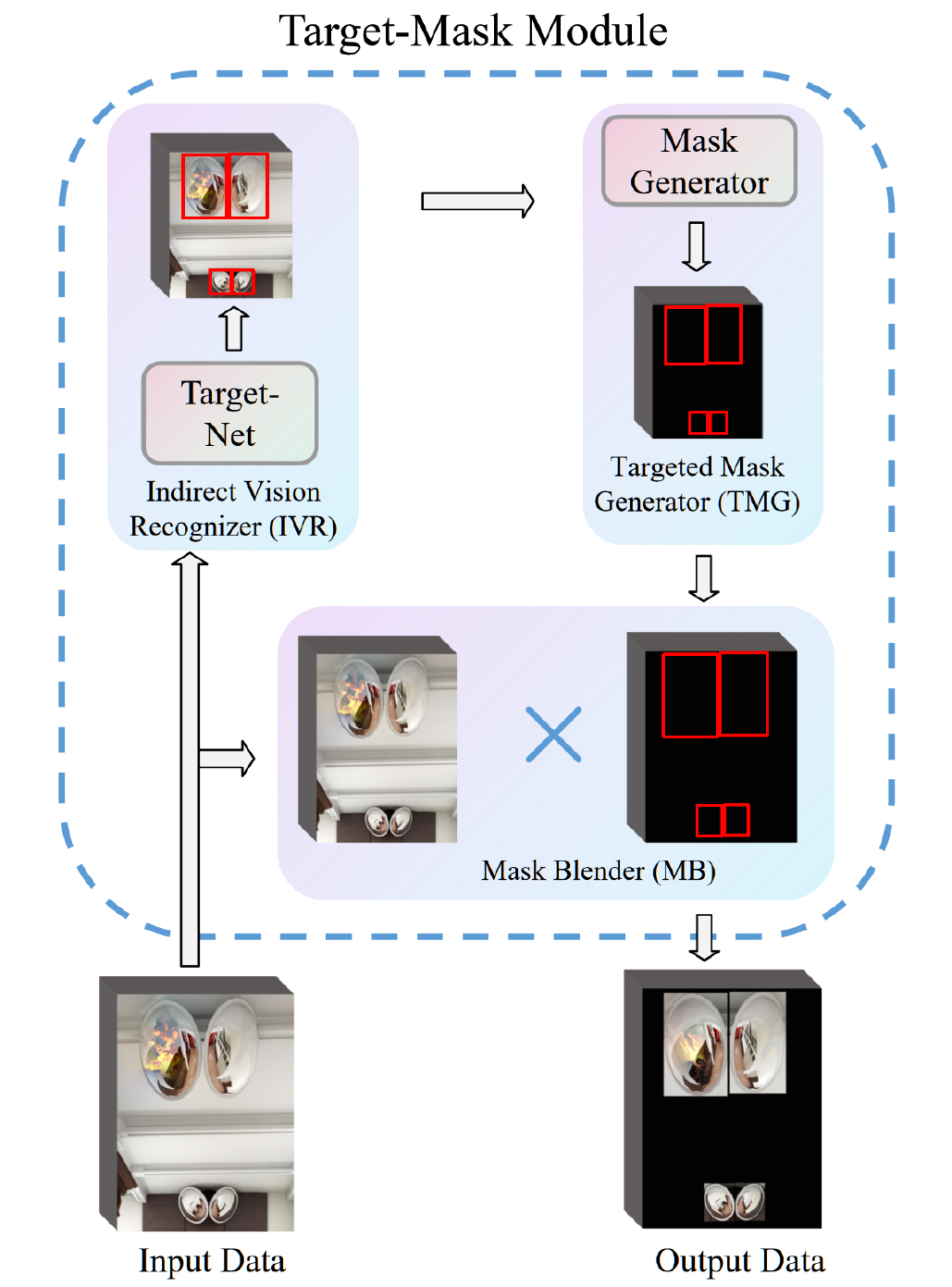}
    \caption{The Target-Mask is added to the neck between the data input and the backbone network of YOLOv8. When the image data passes through the Target-Mask module, the module will use the built-in network to find the area location information of each indirect vision in the image, and then use the obtained location information to generate a mask with corresponding area boundary information. Then, the mask is mapped back to the original image. After the mapping processing, the passed image only retains the pixels of the target area and filters out other non-interest areas. Then, the optimized image data is sent to the subsequent network of YOLOv8.}
    \label{fig:target-mask}
\end{figure}

In the fire monitoring tasks of heritage buildings, in order to reduce drilling and wiring, we will utilize the indirect field of view to expand the coverage area of a monocular camera, thereby reducing the number of cameras required and simultaneously achieving the alignment of the indirect field of view with the target area. When it is necessary to handle the monitoring tasks by using the indirect field of view, to solve the problems of easy false detections and omissions of the model in this new and complex situation, we have designed the most targeted Target-Mask module and inserted this module into the neck part between the image input and the backbone network. As shown in Figure \ref{fig:target-mask}, the Target-Mask module contains three sub-modules: Indirect Vision Recognizer (IVR), Targeted Mask Generator (TMG), and Mask Blender (MB). The IVR includes an indirect vision targeting network called Target-Net. The weights of the built-in network of Target-Net are transferred through pre-learning. During the first round of training, the Target-Net identifies the indirect field of view in the image and then transmits the read position information to the TMG. Based on the obtained position information, the TMG will generate a mask for the targeted area and then pass it to the MB. After receiving the targeted mask, the MB will fuse and superimpose the targeted mask with the passing image data, so that the passing image will only retain the image within the indirect field of view area. In this way, when the subsequent image passes through the Target-Mask module, the generated mask will be mapped onto the passing image, and the image will only retain the area of the indirect field of view, while other areas will be filtered out as non-interest areas. This can ensure that only the indirect field of view exists in the detected image, preventing confusion between the indirect and direct fields of view. Meanwhile, through this kind of partitioning and filtering, the model can focus more on the indirect field of view area and eliminate the interference from irrelevant areas. This targeted detection helps the model accurately identify the fire characteristics within the target area under complex environmental backgrounds. Meanwhile, compared with the traditional CBAM and CA, the Target-Mask is a brand-new module based on the indirect field of view detection task. It not only has higher performance but is also more efficient in terms of the required parameters. Therefore, this paper chooses to introduce the Target-Mask module. By introducing the Target-Mask module into the neck part of YOLOv8, the model pays more attention to the position information of the target area to improve the detection accuracy of the target area and reduce false alarms.

\section{ Experiment and Analysis} 

\subsection{Datasets}

\begin{figure}[t]
    \centering
    \includegraphics[width=\linewidth]{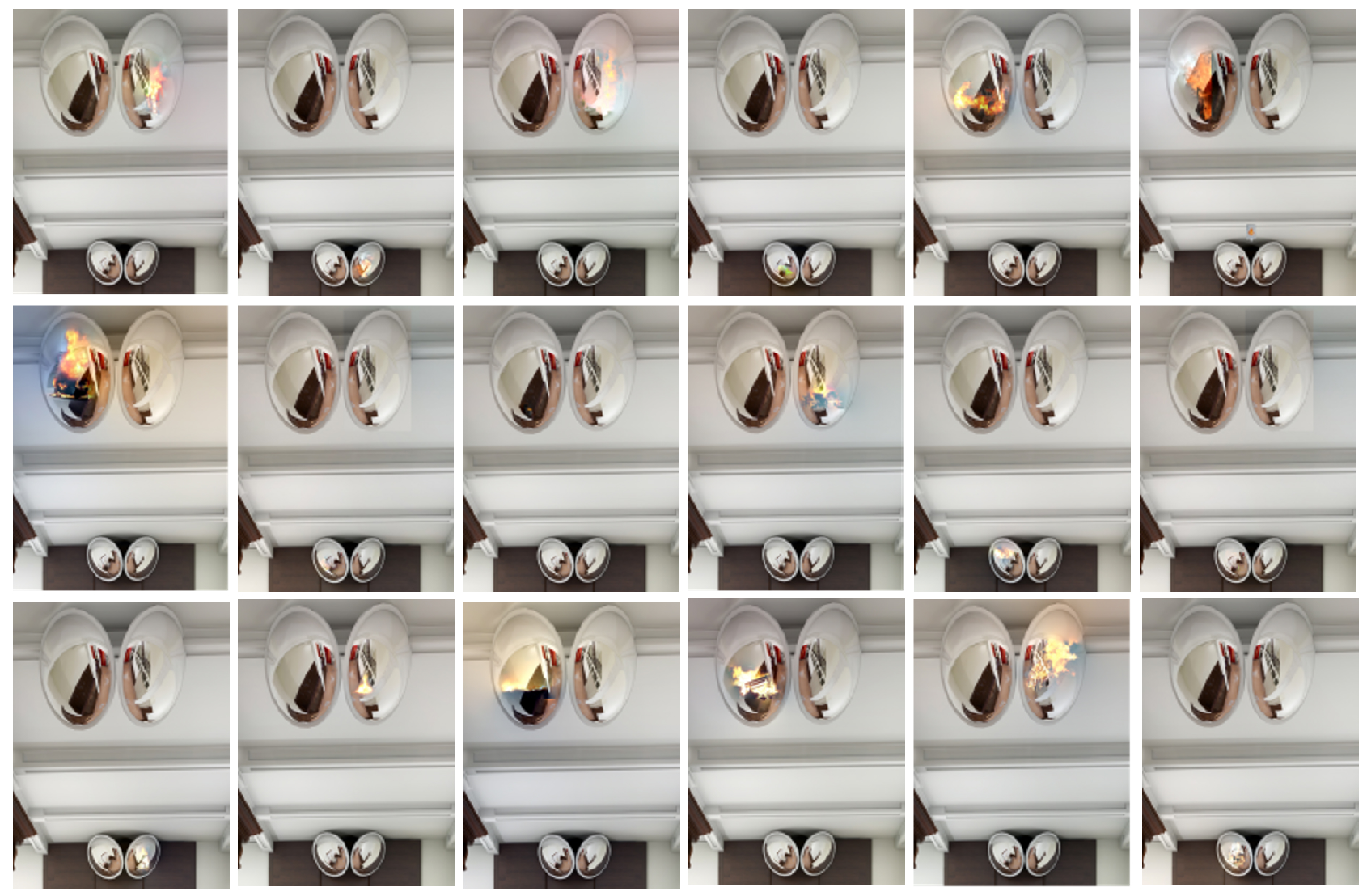}
    \caption{Example of experimental data.}
    \label{fig:firedata}
\end{figure}

We take the former site of the Military Affairs Committee of the Guangdong District Party Committee of the Communist Party of China located in China as the research object. It happens to be a provincial-level cultural relic protection unit that is troubled by fire prevention monitoring. Due to the particularity of cultural relic buildings, it is impossible to conduct actual fire tests inside, which is also not allowed by relevant cultural relic protection laws and regulations. Therefore, we use 3DMAX to conduct virtual fire scenes for it for subsequent fire detection experiments.
By setting and adjusting the wide-angle mirror in the model and then using 3DMAX's rendering method, we can obtain the corresponding scene image. In this experiment, we set up four target monitoring areas and one non-interest area. Through adjustment, the indirect vision in the mirror has been aligned with the target monitoring area. At the same time, due to the ceiling position of the venue, there is no line passing through. The ceiling is made of non-combustible concrete material and is also far away from the crowd. The probability of a fire occurring first in this area tends to zero. Therefore, we do not set up a wide-angle mirror to observe this area, that is, this area is not included in the indirect vision and is used as a non-interest area. In this non-interest area, in a small amount of data, we will hang some flags as some noise in the detection to test the anti-interference ability of the detection model. This is also a real situation in the actual operation of the venue. Then, we collect images with the camera facing the main entrance of the building to form this fire data set based on indirect vision. This data set contains a total of 800 fire pictures and is saved in JPG format with a pixel size of 1645×2493; at the same time, in order to verify the model's ability to resist noise interference in non-interest areas, 100 of them have added flags as noise to the non-interest areas. Finally, they are randomly divided into a training set of 560 images, a validation set of 120 images, and a test set of 120 images. Some images of this data set are shown in Figure \ref{fig:firedata}.

\subsection{Experimental Environment and Parameter Setting}

The experimental operating system used in this study is Win 10, while PyTorch serves as the underlying framework for the developed deep learning model. The detailed environment is shown in Table \ref{tab:t1}. The hyperparameters of the model training phase include input image size is 640 × 640, the batch size is 64, the optimizer is SGD stochastic gradient descent, and the number of training rounds is 100 epochs. The learning rate is initialized to 0.01, with momentum and weight decay values set at 0.937 and 0.0005, respectively. All other training parameters use the default values of the YOLOv8n model. In addition, the official pretraining weight files provided in this study were used to enhance the generalization of the trained model.

\begin{table}[htbp]
\centering
\caption{Training environment and hardware platform parameters table.}
\begin{tabular}{cc} 
\toprule 
Parameters  & Configuration   \\
\midrule 
CPU & I9-12490F  \\
GPU & NVIDIA GeForce RTX 4060 \\
GPU memory size & 8G  \\
Operating systems & Win 10  \\
Deep learning architecture & Pytorch1.9.2 + Cuda11.4 + cudnn11.4  \\
simulation software & 3DMAX  \\
\bottomrule 
\end{tabular}
\label{tab:t1}
\end{table}

\subsection{Evaluation Metrics} 
The main metrics used in this paper are Precision, Recall, and mAP. Precision refers to the proportion of samples that are actually positive in all the samples predicted by the model as positive categories, which measures the accuracy of the model in the prediction of positive categories; Recall refers to the proportion of samples that are actually positive in all the samples that are correctly predicted by the model as positive categories, which measures whether the model is able to efficiently find all the positive categories of samples; The mAP is a commonly used evaluation metric in target detection tasks that combines the precision and recall curves of the model on different categories and calculates the average value, which measures the detection performance of the model on multiple categories, and is often used to evaluate the overall effectiveness of target detection algorithms. Where mAP50 denotes the mAP value at the 50\% loU threshold. 

\subsection{Comparative Experiments on Different Attention Mechanisms
}

Table \ref{tab:attention} presents the outcomes of integrating different attention mechanisms, encompassing four distinct types: CBAM, CA, SA~\cite{zhang2021sa}and Target-Mask. YOLOv8 incorporating the Target-Mask module yields the most advantageous results. Specifically, in comparison to the original YOLOv8, it exhibits a significant improvement of 6.7\% in Precision. Additionally, in the comparison of Precision indicators, Target-Mask outperforms the CBAM module by 5\%. This indicates that in scenarios where detection is conducted using indirect vision, Target-Mask has the greatest enhancement in detection performance compared to other attention modules. This is primarily attributed to the fact that Target-Mask directly filters out the interference from non-interest areas, effectively reducing the false alarm rate of the model. In the fire prevention monitoring task of cultural relic buildings, high Precision can reduce the frequency of false alarms, thereby avoiding unnecessary waste of human resources and on-site panic.

\begin{table}[t]
\centering
\caption{Comparative experiments on attention mechanisms.}
\begin{tabular}{l@{\hspace{2.5cm}}c@{\hspace{2.5cm}}c}
\toprule 
Model & P(\%) & mAP50(\%) \\
\midrule 
YOLOv8 & 87.1 & 87.9 \\
+CBAM~\cite{CBMA2018} & 87.8 & 88.6 \\
+CA~\cite{CA2021} & 89.2 & 88.5 \\
+SA~\cite{zhang2021sa} & 88.9 & 88.9 \\
+Taget-Mask & 93.8 & 91.6 \\
\bottomrule 
\end{tabular}
\label{tab:attention}
\end{table}

\subsection{Ablation Experiment}
\begin{figure}[t]
    \centering
    \includegraphics[width=\linewidth]{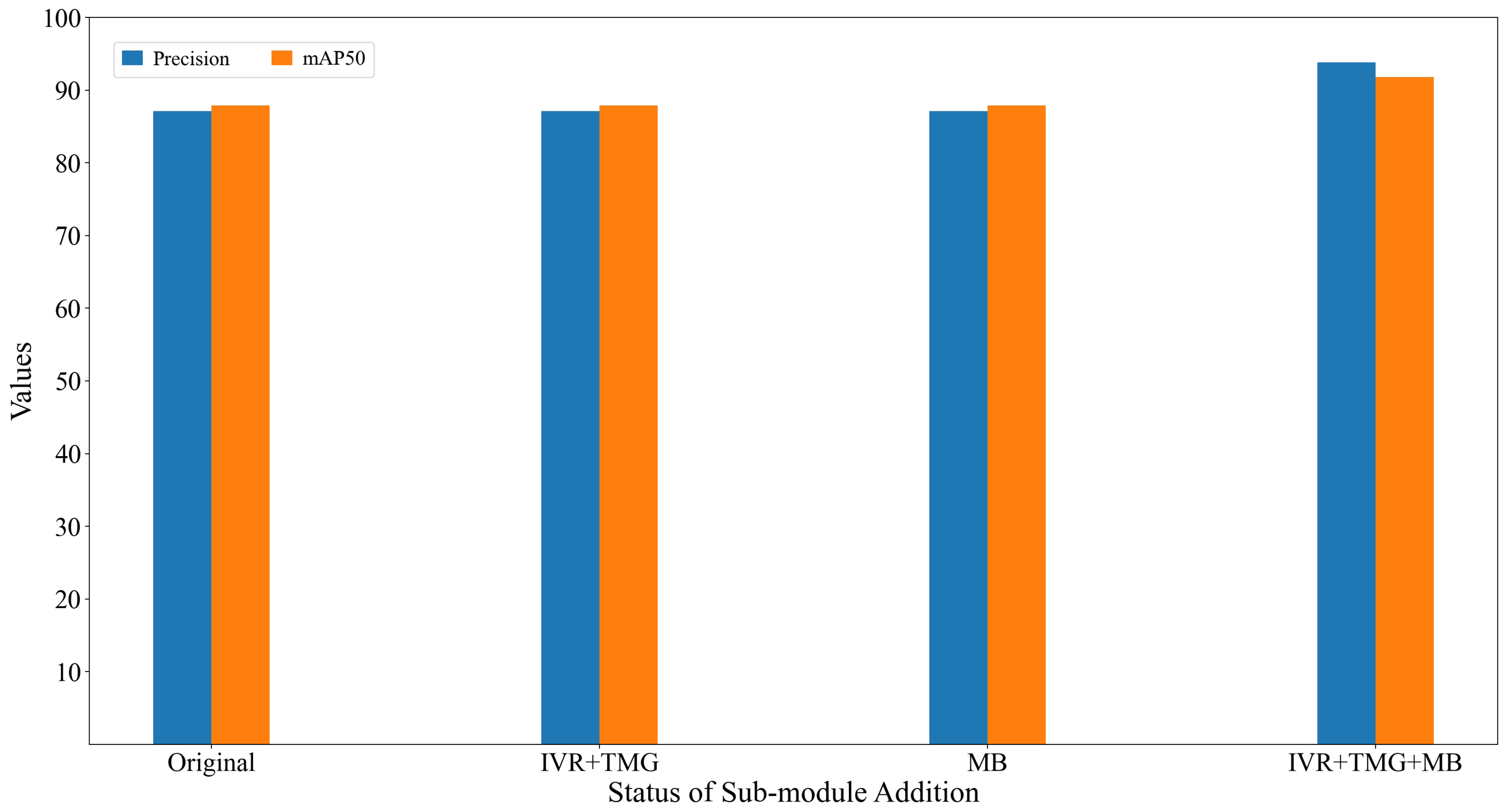}
    \caption{Results of ablation experiment.}
    \label{fig:histogram2}
\end{figure}

To verify the functions and mutual influences of each sub-module in Target-Mask, this ablation experiment was carried out under the same dataset and training parameters. The results are shown in Table \ref{tab:Ablation} and Figure \ref{fig:histogram2}. In the case of not adding any sub-modules, it represents the normal detection performance of the baseline model YOLOv8n. When only the MB sub-module is removed among the three sub-modules in Target-Mask, although the TMG generates the targeted mask, without the MB sub-module to superimpose and filter the generated mask with the original data, the original image passes through smoothly, and the detection performance is the same as that of the baseline model YOLOv8n. When only the TMG sub-module is removed, it means that no mask will be generated and passed to the MB module anymore, so the original data will also pass through smoothly without being processed. However, when the IVR sub-module is removed, since the TMG doesn't obtain the position data of the indirect field of view, it will assume that there is no indirect field of view in the original image and thus generate a completely black mask, which is passed to the MB sub-module to be superimposed with the original image data. This will filter out the entire image. It can be seen that when the three sub-modules exist simultaneously, the performance of the model is improved and enhanced. Therefore, from the experimental results, it can be known that the three sub-modules of Target-Mask are interdependent. If any one of them is removed, the desired effect will not be achieved.

\begin{table}[t]
\centering
\caption{Results of ablation experiment.}
\begin{tabular}{l@{\hspace{12mm}}c@{\hspace{12mm}}c@{\hspace{12mm}}c@{\hspace{12mm}}c}
\toprule 
IVR & TMG & MB & P(\%) & mAP50(\%) \\
\midrule 
 -- &  --  &  -- & 87.1 & 87.9 \\
\text{\checkmark}  & \text{\checkmark} & --  & 87.1 & 87.9 \\
\text{\checkmark}  & -- &\text{\checkmark} & 87.1 & 87.9 \\
\text{\checkmark}& \text{\checkmark} &\text{\checkmark} & 93.8 & 91.6 \\
\bottomrule 
\end{tabular}
\label{tab:Ablation}
\end{table}

\subsection{Comparison with Other Models}

\begin{figure}[t]
    \centering
    \includegraphics[width=\linewidth]{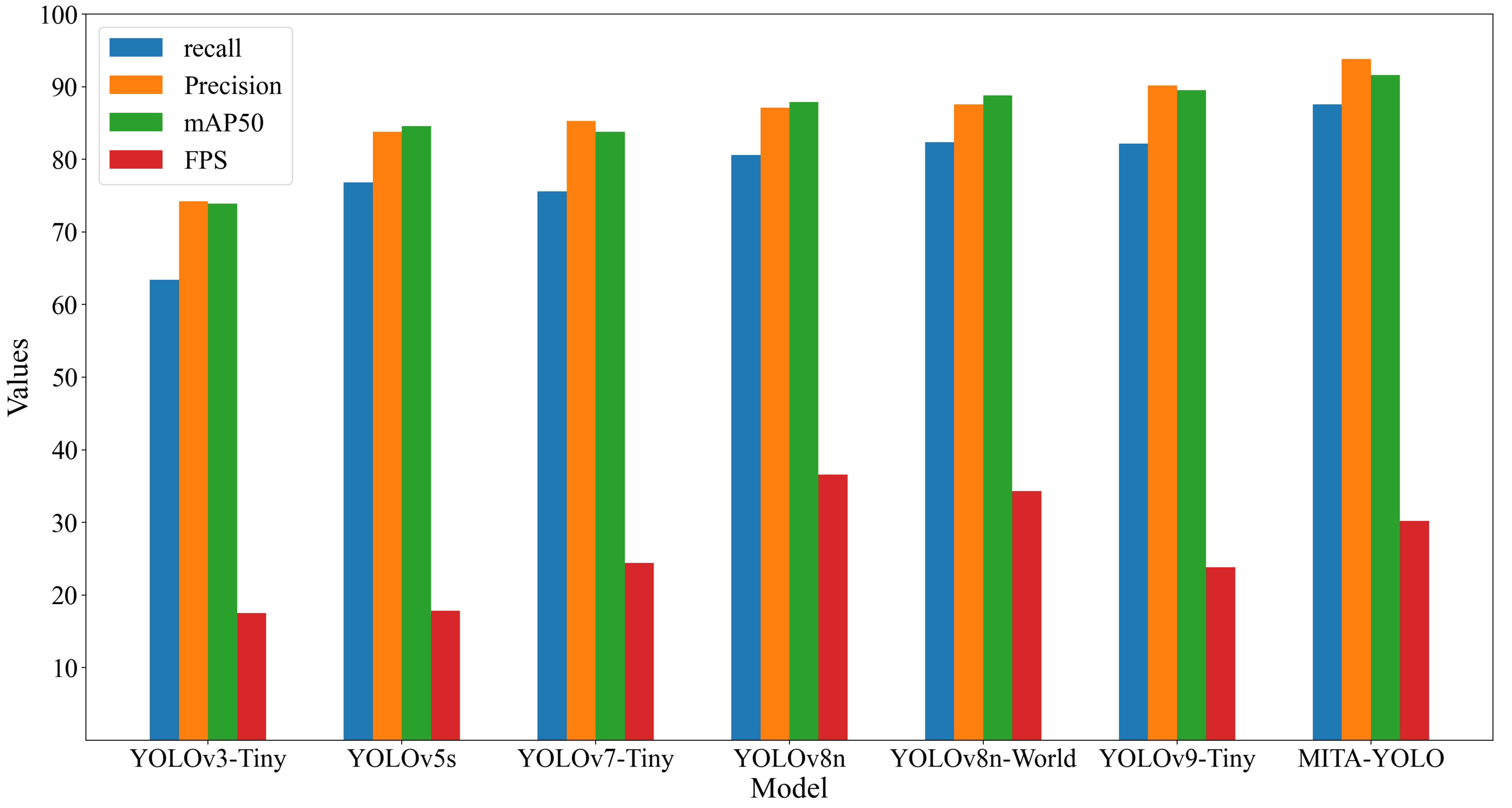}
    \caption{Results of comparative experiments.}
    \label{fig:histogram}
\end{figure}

To comprehensively investigate the performance of the improved model proposed in this paper in the fire detection task, a series of comparison experiments were carried out with other renowned target detection models under identical dataset and training parameters. These models comprise YOLOv3-Tiny~\cite{fu2021fast}, YOLOv5s~\cite{wu2021application},YOLOv7-Tiny~\cite{yang2023tea}, YOLOv8n~\cite{lou2023dc}, YOLOv8n-World~\cite{zhang2024research}, and the most recent YOLOv9-Tiny~\cite{wang2025yolov9}. The results of these comparison experiments are presented in Table \ref{tab:othermodels} and Figure \ref{fig:histogram}. Evidently, in comparison to other models, the improved model attains the highest values for the three evaluation metrics of recall rate, precision rate, and mAP50, which robustly validates the effectiveness of model improvement. When comparing Target-YOLO with the baseline model YOLOv8n, it is observed that the mAP50 value of the improved model is increased by 3.7\%. This significant increase clearly indicates that the performance of the improved model in determining the target position and classifying the target has been effectively enhanced. Simultaneously, the recall rate of the improved model is increased by 3\%. The elevation in recall rate implies that the model's capability to capture important target objects has been augmented. In the fire monitoring of cultural relic buildings, the omission of fire objects or abnormal events may lead to grave consequences. However, a model with a high recall rate can significantly reduce this risk. Through the Target-Mask module, the model can completely focus on detecting the target area, which greatly enhances the image feature extraction ability of key areas and the anti-interference ability with respect to non-interest areas. Consequently, the model is able to better capture the feature information of target objects, thereby upgrading the overall performance of the model.

\begin{table}[t]
\centering
\caption{Results of comparative experiments.}
\begin{tabular}{l@{\hspace{7mm}}c@{\hspace{7mm}}c@{\hspace{7mm}}c@{\hspace{7mm}}c}
\toprule 
Model&recall& P(\%) & mAP50(\%) &	FPS   \\
\midrule 
YOLOv3-Tiny~\cite{fu2021fast}    &73.4	&74.2   &73.9	&17.5 \\
YOLOv5s~\cite{wu2021application}	       &82.8	&83.8   &84.6    &17.8 \\
YOLOv7-Tiny~\cite{yang2023tea}	   &79.6    &85.3	&83.8    &24.4 \\
YOLOv8n~\cite{lou2023dc}        &84.6	&87.1   &87.9	&36.6 \\
YOLOv8n-World~\cite{zhang2024research}	&85.4   &87.6   &88.8&	34.3 \\
YOLOv9-Tiny~\cite{wang2025yolov9}	   &85.2    &90.2	&89.5   &23.8 \\
MITA-YOLO  &87.6    &93.8	&91.6	&30.2 \\
\bottomrule 
\end{tabular}
\label{tab:othermodels}
\end{table}

\subsection{Experimental Effect Verification}

\begin{figure}[t]
    \centering
    \includegraphics[width=\linewidth]{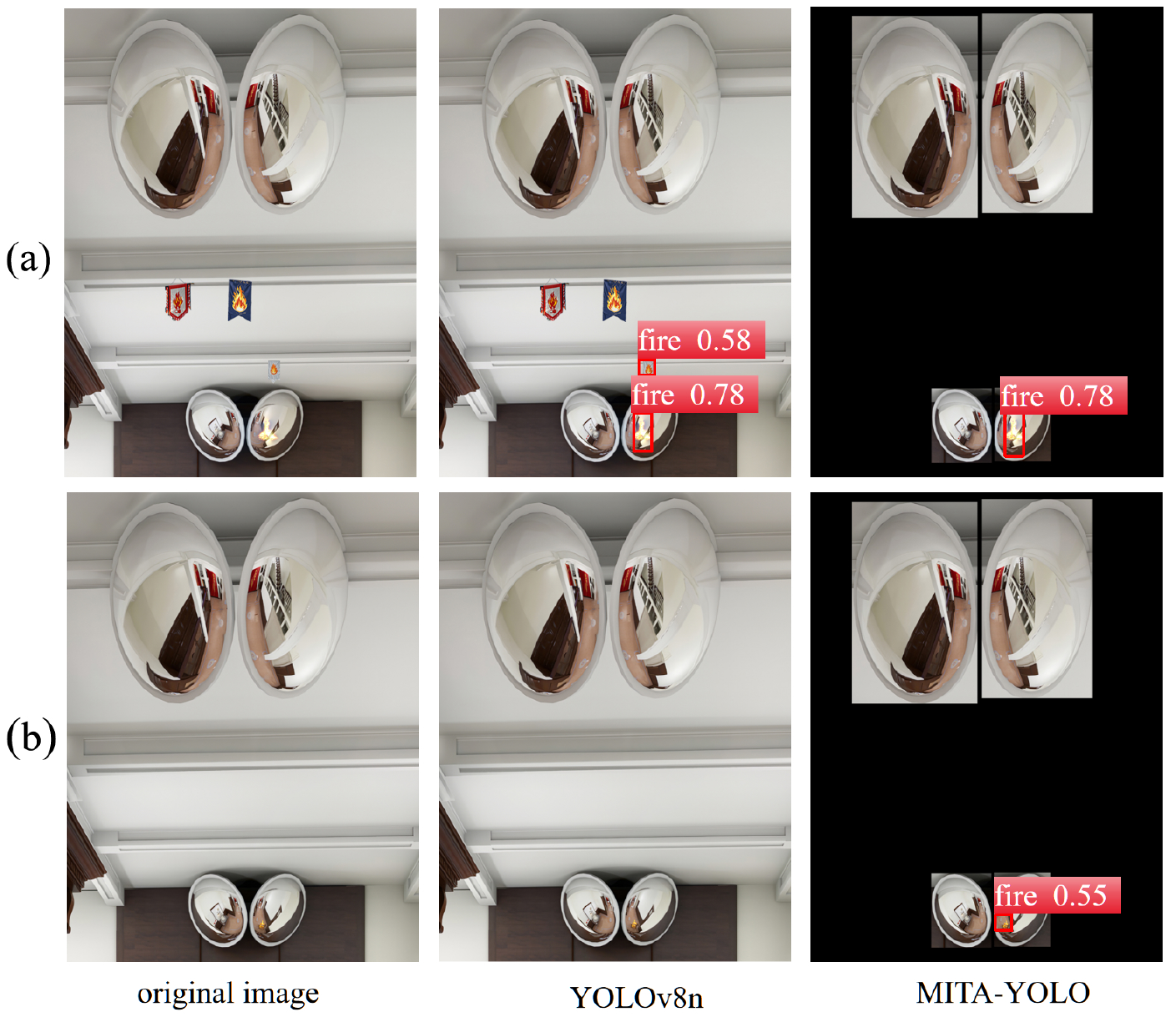}
    \caption{Comparison of detection results between YOLOv8n and MITA-YOLO.}
    \label{fig:det-text}
\end{figure}

In Figure \ref{fig:det-text}, we select some detection result samples from the test set to clearly show the differences brought about by model improvement. The left column is the original image obtained by the camera, the middle is the detection result using the YOLOv8n model, and the right column shows the detection result of the MITA-YOLO model. First, in the detection of data a by YOLOv8n, it can recognize the fire target in the indirect vision in the lower right corner. At the same time, it can also recognize that the images in the two flags in front of the non-interest area cannot be used as fire targets. However, the pattern in the white flag at the rear is mistakenly identified as a fire target; while the detection of data a by MITA-YOLO, in addition to being able to recognize the fire target in the indirect vision in the lower right corner, since the non-interest area is directly filtered out, it will not be affected by the noise in the non-interest area at all and generate false alarms. Secondly, in the detection of data b by YOLOv8n, the small fire target in the indirect vision in the lower right corner is missed; while in the detection of data b by MITA-YOLO, since only the indirect vision is used as the targeted area, the concentration is stronger, so that the small fire target in the indirect vision in the lower right corner is successfully recognized.

From the above comparison results, we can know that although YOLOv8 can understand through training that the patterns in the flag should not be recognized as fire targets, false alarms will still occur when some unfamiliar flag shapes or positions change. The detection area of MITA-YOLO is targeted, and it can directly inherit the experience of artificial priori to remove the interference of irrelevant areas, greatly reducing the false alarm rate and bringing great convenience to the actual work of the venue. On the other hand, since MITA-YOLO only focuses on detecting indirect vision areas and pays greater attention to the targets in the indirect vision area, in the scenario based on indirect vision, its overall detection accuracy is higher and the missed detection rate is also greatly reduced.

\section{Conclusion}\label{sec:con} 

In this study, we constructed a fire detection method based on indirect vision, namely MITA-YOLO. The innovation of this method resided in proposing the transfer of the advantages of indirect vision to the field of deep learning. Firstly, through the judicious arrangement of indirect vision, not only was the monitoring range of a single camera expanded, but also alignment with the target monitoring area was achieved. Then, by employing the Target-Mask module we designed, the model could automatically identify each indirect vision as the targeted detection area and simultaneously filter out other non-interest areas. This design not only significantly reduced the number of cameras and mitigated the damage to the structure of cultural relic buildings and the impact on historical features; moreover, the model could fully inherit the manager's judgment on the area where a fire was likely to occur first, thereby enhancing the concentration and anti-interference ability of the fire detection model. Our method not only improved the fire protection of cultural relic buildings but also diminished the impact of false alarms on venue operations. Based on the indirect vision fire data set we created, when compared with the original YOLOv8 and other mainstream detection models, the experimental results demonstrated that MITA-YOLO exhibited higher accuracy and robustness in the detection task of indirect vision. In addition to being applicable to fire detection of cultural relic buildings, this detection method could also be utilized in various projects that required the expansion of monocular vision coverage, had prior experience in area division that needed to be inherited, or had cost-saving demands.

In the field of deep learning in the future, the way to better combine indirect vision for computer vision tasks still needed further improvement. It was hoped that by continuously improving the generator of indirect vision, expanding more data sets, and designing detection modules with higher compatibility, it could be applied in more complex scenarios. In the long run, just as the automotive industry had widely employed the indirect vision of rearview mirrors to generate tremendous application value, the application potential of indirect vision and its paired detection model in deep learning should be continuously explored and tapped, with the hope of addressing task requirements in more diverse scenarios.

\bibliographystyle{IEEEtran}
{\small
\bibliography{ref}
}
\end{document}